\documentclass[journal,transmag]{IEEEtran}
\usepackage{ifpdf}
\usepackage{cite}
\ifCLASSINFOpdf
\else 
\fi
\usepackage{graphicx}
\usepackage{amsmath}
\usepackage{algorithmic}
\usepackage[flushleft]{threeparttable}
\usepackage{array}
\ifCLASSOPTIONcompsoc
\usepackage[caption=false,font=normalsize,labelfont=sf,textfont=sf]{subfig}
\else
\usepackage[caption=false,font=footnotesize]{subfig}
\fi
\usepackage{url}
\usepackage{multirow}
\begin{document}
\title{BERT based classification system for detecting rumours on Twitter}

\author{Rini~Anggrainingsih \IEEEauthorrefmark{1} \IEEEauthorrefmark{2},
        Ghulam~Mubashar~Hassan \IEEEauthorrefmark{2}
        and~Amitava~Datta \IEEEauthorrefmark{2}}
\thanks{\IEEEauthorrefmark{1} Informatics Department of Mathematics and Natural Sciences Faculty, Sebelas Maret University, Surakarta, Indonesia.}
\thanks{\IEEEauthorrefmark{2}School of Computer Science and Software Engineering, The University of Western Australia, Perth, WA, Australia.}

%
%


\maketitle

\begin{abstract}
The role of social media in opinion formation has far reaching implications in all spheres of society. Though social media provide  platforms for expressing news and views, it is hard to control the quality of posts due to the sheer volumes of posts on platforms like Twitter and Facebook. Misinformation and rumours have lasting effects on society, as they tend to influence people's opinions and also may motivate people to act irrationally. It is therefore very important to detect and remove rumours from these platforms. The only way to prevent the spread of rumors is through automatic detection and classification of social media posts. Our focus in this paper is the Twitter social medium, as it is relatively easy to collect data from Twitter. The majority of previous studies used supervised learning approaches to classify rumours on Twitter. These approaches rely on feature extraction to obtain both content and context features from the text of tweets to distinguish rumours and non-rumours. Some researchers reported that tweets contain a linguistic pattern that can be used as a crucial features in identifying rumours on Twitter. Manually extracting features however is time consuming considering the volume of tweets. We propose a novel approach to deal with this problem by utilising sentence embedding using BERT to identify rumours on Twitter, rather than the usual feature extraction techniques.  BERT is a novel transformer that can capture the contextual meaning of words.  We use sentence embedding using BERT to represent each tweet’s sentences into a vector according to the contextual meaning of the tweet. We classify those vectors into rumours or non-rumours by using various supervised learning techniques. Our BERT based models improved the accuracy by approximately 10\%  as compared to previous methods.  Our best model uses a multi-layer perceptron and achieves 0.869 accuracy, 0.855 precision, 0.848 recall and 0.852 F1 score. 
\end{abstract}

\begin{IEEEkeywords}
BERT, Rumour Detection, Sentence Embedding, Transformer, Twitter
\end{IEEEkeywords}

%
\IEEEpeerreviewmaketitle

\section{Introduction}
The lack of control over the posts on social media platforms such as Twitter and Facebook results in the rapid spread of false information. People tend to post and share breaking news without verifying the credibility of the content, and thus, even though it may have false contents, a piece of  information may be shared thousands of times before its contents can be verified. \textit{Rumours} look credible on the surface, yet they are unverified and often false news items. Rumours circulate with questionable trustworthiness and cause anxiety  and irrational behaviour among the readers \cite{Zubiaga2016, pamungkas2019stance}. Rumours could be true, partly true, false or may remain unverified \cite{Bondielli2019}. In this study, we focus on the spread of rumours via the Twitter platform.

Countless rumours on different topics spread on Twitter every day and influence people's opinions and decisions. Rumours also lead to harmful effects on society. For example, the tweet "Explosion at White House and Obama injured" shared on Twitter in 2013 resulted in stock markets crashing and investors lost around 136 billion dollars within only two minutes \cite{Keller2013}. Hence, there is a need to detect the trustworthiness of shared information on Twitter and detect rumours automatically since manual fact-checking is a time consuming process and a highly non-trivial task. 

 Most of the existing studies on rumour detection on Twitter  rely on reliable feature extraction. These studies use supervised learning approaches to classify the  authenticity of tweets by manually extracting features from the context and content of tweets  \cite{Bondielli2019,Castillo2011,ito2015assessment,Zubiaga2016,Hassan2018}. The context-based feature components involve tweets' associated information, such as users' information and network information \cite{Castillo2011,ito2015assessment,Zubiaga2016,Hassan2018,Ghenai2017,herzallah2018feature}. Whereas the content-based feature components involve extraction of features from the text of tweets, particularly the linguistic aspects such as lexical, syntactic and semantic information \cite{Castillo2011,ito2015assessment,Zubiaga2016,Hassan2018,Ghenai2017,herzallah2018feature,sato2018credibility,kotteti2018multiple}. Feature extraction however often needs additional information which is not always available on Twitter \cite{Ma2015}.
 
 Recently, deep learning methods have gained popularity in detecting rumours on tweets. These methods have the ability to discover features automatically and exploit the deep data representation from the text, which is beneficial for efficient rumour detection. These approaches include recurrent neural network (RNN) \cite{Ma2015,Ruchansky2017}, convolutional neural network (CNN) \cite{Yu2017,Bharti2021} and long short term memory (LSTM) networks \cite{Ajao2018,Alkhodair2020}. Some studies proposed the use of reinforcement learning \cite{Zhou2019} to improve rumour detection. However, the performance of deep learning models need to be improved for automatic rumour detection to be a practical method.

 We propose a novel approach using sentence embedding of Bidirectional Encoder Representation from Transformer (BERT) to detect whether a tweet is a rumour or not. BERT is a novel transformer technique that can capture contextual meanings of a word by considering words from both left and right side of it \cite{devlin2018bert}. We use BERT to represent each tweet into a vector based on the contextual meaning of its sentence and use different classifiers to detect rumours on Twitter. We found that our proposed technique achieves better results than existing state-of-the-art techniques. 

The rest of the paper is organized as follows. Section 2 discusses related works on rumour detection. Section 3 presents the details of our proposed approach on utilising BERT to detect rumours in tweets. In Section 4, we present experimental results from our methods and comparison with state-of-the-art results. We conclude the study in Section 5. 

\section{Related Works}

\begin{figure}[hbt!]
    \centering
    \includegraphics [scale=0.8]{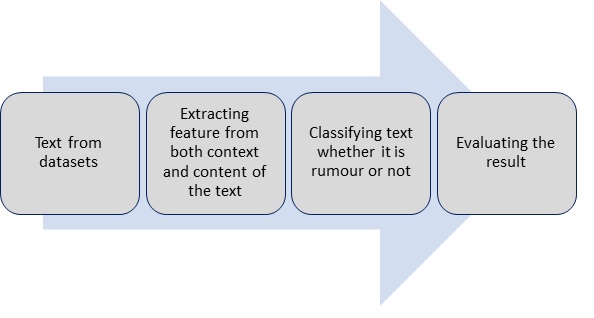}
    \caption{General architecture of rumour classification models.}
    \label{fig:Figure 1}
\end{figure}

The majority of existing studies on rumour detection used supervised learning models to classify tweets into trustworthy and untrustworthy, are based on extracted context and content features \cite{Castillo2011,ito2015assessment,Ghenai2017,herzallah2018feature,Hassan2018}. Figure \ref{fig:Figure 1} illustrates the general steps involved in these techniques. Context-based approaches extract features by taking into account information of tweets including user and network information. Table \ref{table1} summarises context-based features which are extracted from tweets in previous methods.

\begin{table}[ht]
\caption{Context-based features extracted from the tweets.}
\begin{tabular} {|p{4.5cm}|p{3cm}|} 
\hline
Context-based features & References \\
\hline

It is verified account & \cite{Castillo2011}, \cite{ito2015assessment}, \cite{Hassan2018}  \\
Account has a description & \cite{Castillo2011}, \cite{Hassan2018}  \\
Account has a URL & \cite{Castillo2011}, \cite{ito2015assessment}, \cite{Hassan2018} \\
The number of followers & \cite{Castillo2011}, \cite{ito2015assessment}, \cite{Hassan2018}, \cite{herzallah2018feature}, \cite{Ghenai2017}, \cite{chatterjee2018classifying} \\
The number of friends & \cite{Castillo2011}, \cite{ito2015assessment}, \cite{Hassan2018}, \cite{herzallah2018feature}\\
Account has over 500 followers & \cite{chatterjee2018classifying} \\
Post on a day or weekday & \cite{Castillo2011}, \cite{Ghenai2017}\\
The number of posts & \cite{Castillo2011}, \cite{herzallah2018feature}, \cite{Ghenai2017} \\
Is it retweeted or not & \cite{Castillo2011}, \cite{Hassan2018}, \cite{Ghenai2017}\\
\hline
\end{tabular}
\label{table1}
\end{table}

Content-based approaches extract features from tweets, in particular the linguistic aspects including lexical characteristics, e.g., doubt words, negation, clear expression and abbreviation, which can indicate a tweet's trustworthiness \cite{Bondielli2019}. These features utilise  Natural Language Processing (NLP) techniques to capture opinion and emotional expressions in a tweet. Table \ref{table2} illustrates the content-based features and the studies which used them.

\begin{table}
\caption{Content-based features extracted from the tweets.}
\centering
\begin{tabular}{|p{4.5cm}|p{2.8cm}|} 
\hline
Content-based features & References \\
\hline
The number of hashtags & \cite{Castillo2011}, \cite{ito2015assessment}, \cite{Ghenai2017}, \cite{herzallah2018feature}, \cite{Hassan2018}\\
Words length & \cite{Castillo2011}, \cite{ito2015assessment}, \cite{Ghenai2017}, \cite{herzallah2018feature}, \cite{Hassan2018}\\
Characters length & \cite{Castillo2011}, \cite{ito2015assessment}, \cite{Ghenai2017}, \cite{Hassan2018}\\
Contains 100 top domain & \cite{Castillo2011} \\
Is it contains URL & \cite{Castillo2011} \cite{ito2015assessment} \cite{Hassan2018}\\
The number of URLs & \cite{Castillo2011}, \cite{ito2015assessment}, \cite{Hassan2018}\\
Mention news agency & \cite{chatterjee2018classifying}\\
The number of mention users & \cite{Castillo2011}, \cite{ito2015assessment}, \cite{Ghenai2017}, \cite{Hassan2018}\\
Contains stock symbol & \cite{Castillo2011}\\
Contains numbers & \cite{chatterjee2018classifying}\\
Contains selected users & \cite{Castillo2011}\\
The number of uppercase & \cite{Castillo2011}, \cite{Ghenai2017}\\
Number of question mark & \cite{Castillo2011}, \cite{ito2015assessment}, \cite{Ghenai2017}, \cite{Hassan2018}\\
Number of exclamation mark & \cite{Castillo2011}, \cite{ito2015assessment}, \cite{Ghenai2017}, \cite{Hassan2018}\\
Contains multi '?' or '!' & \cite{Castillo2011}, \cite{ito2015assessment}, \cite{Ghenai2017}\\
The number of smile emote & \cite{Castillo2011}, \cite{Ghenai2017}, \cite{Hassan2018}\\
The number of frown emote & \cite{Castillo2011},  \cite{Ghenai2017}, \cite{Hassan2018}\\
Number of sentiment (+) words & \cite{Castillo2011}, \cite{Ghenai2017} \\
Number of sentiment (-) words & \cite{Castillo2011}, \cite{Ghenai2017} \\
Sentiment score & \cite{Castillo2011}, \cite{Ghenai2017} \\
The number of 1st pronouns & \cite{Castillo2011}, \cite{Ghenai2017}, \cite{hu2013dude}\\ 
The number of 2nd pronouns & \cite{Castillo2011}, \cite{Ghenai2017}, \cite{hu2013dude}\\ 
The number of 3rd pronouns & \cite{Castillo2011}, \cite{Ghenai2017}, \cite{hu2013dude}\\
The number of temporal reference & \cite{hu2013dude}\\
The number of lexical density & \cite{hu2013dude}\\
The number of Terminology (slang) & \cite{chatterjee2018classifying}\\
The number of intensifiers & \cite{chatterjee2018classifying}\\
Contains repeated characters & \cite{chatterjee2018classifying}\\
Contains all uppercase word & \cite{chatterjee2018classifying}\\
Title capitalisation & \cite{chatterjee2018classifying}\\
\hline
\end{tabular}
\label{table2}
\end{table}

\par The features  mentioned in Table \ref{table1} and Table \ref{table2}, are usually  manually hand-crafted features either contextual information or information from the texts of tweets. Feature extraction  (either content-based or context-based) is a very time-consuming and almost an impossible process, considering that hundreds of millions of tweets are generated every day. This led to the use of Neural Network (NN) based techniques to classify rumours in tweets. In the context of rumour detection on Twitter, the widely implemented NN based frameworks are Recurrent Neural Network (RNN) \cite{Ma2015,Ruchansky2017,Alkhodair2020} and Convolutional Neural Network (CNN) \cite{Yu2017,Bharti2021}. The recent performance of RNN models for rumour detection reported by Alkhodair {\em et. al.} achieved F1 scores of 0.716 and 0.839  for rumour and non-rumour classes respectively \cite{Alkhodair2020}.  The latest CNN model for rumour classification was presented by Bharti et al. \cite{Bharti2021} that achieved the highest performance of 0.88 and 0.77 F1-scores for non-rumour and rumour classes respectively. 

 Ajao et al. \cite{Ajao2018} proposed a hybrid framework using CNN and LSTM to automatically identify features from a tweet without prior knowledge of the subject domain or topic of discussion for detecting rumours. They obtained  accuracy and precision of 0.8229  and  0.4435 respectively. Similarly, Kotteti et al. \cite{kotteti2018multiple} attempted to improve supervised learning models for rumour detection by reducing detection time. They proposed a multiple time-series data analysis model to reduce computational complexity by only using the tweets' temporal properties instead of  contents that requires careful feature selections and text mining. Their proposed method reduced the computational complexity significantly and obtained 0.94 precision score by using Gaussian Naïve Bayes classifier. However, they achieved 0.356 for recall and 0.518 for F1-score which are not high.  

In another study, Zhou {\em et al.} \cite{Zhou2019} proposed Early Rumour Detection system (ERD) using reinforcement learning to identify a rumour at early stages. The ERD consists of a rumour detection module and a checkpoint module which determines when to trigger the rumour detection module. They integrated reinforcement learning for the checkpoint module to guide the rumour detection module using classification accuracy as a reward. They treated the incoming tweets as a data stream and monitored the tweets in real time. The tweets are used to determine whether the rumour detection module is triggered. This study reported an accuracy of 0.858.

Xu {\em et al.} \cite{xu2020near} proposed Topic-Driven Novel Detection (TDRD) to determine the credibility of a tweet by its microblog source only. The study mentioned that according to communication theory, a tweet's topic can help to determine the possibility of a tweet to be a rumour or not \cite{allport1947psychology,rosnow1988rumor}. They used two datasets to train their model, Pheme dataset from Twitter and Chinese dataset from Weibo. They utilised a 300-dimensional Glove embedding for Twitter datasets and a 300-dimensional word2vec embedding for Weibo datasets. To classify tweets into rumours, they used CNN and FastText for Twitter and Weibo datasets respectively. The proposed CNN model had two hidden layers while FastText model had 256 hidden layers. The study reported an accuracy of 0.8266.

Some previous studies have reported that semantic and language characteristics are the crucial features to identify rumours \cite{Castillo2011,Zubiaga2016,Ghenai2017,hu2013dude}. 

In this study, we utilise sentence embedding using BERT to extract contextual meaning of tweet's sentences and reveal the specific linguistic patterns of a tweet. To improve the state-of-the-art results, we propose a model using BERT and neural network to classify tweets into rumour and non-rumour tweets. 

\section{Material and Method}

\begin{figure}[hbt!]
    \centering
    \begin{center}
    \includegraphics[scale=0.7]{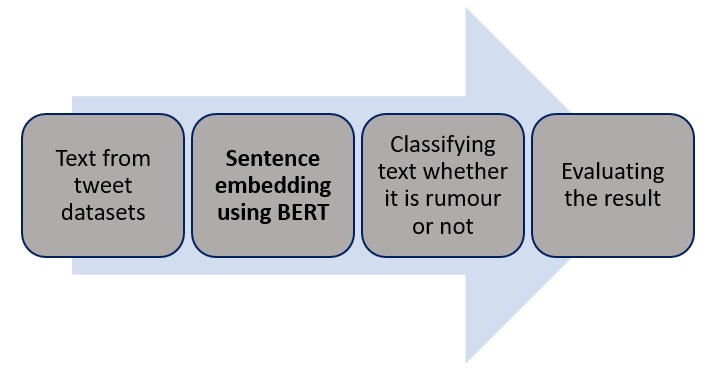}
    \caption{The proposed model for rumour detection using BERT.}
    \label{fig:Figure 2}
    \end{center}
\end{figure}

We propose a  model to classify rumours on Twitter by utilising sentence embedding using BERT instead of feature extraction procedures. Previously, word embedding has been widely used in various natural language processing (NLP) tasks. It is a language representation technique by mapping words or phrases into vectors to use an artificial intelligence model to recognise and perform mathematical operations, and capture a word's semantic importance in a numerical form \cite{mikolov2013distributed}. Besides dealing with individual words, BERT enables working with sentences using sentence embedding. This approach represents the semantic information of the sentences into numerical vectors to help a machine learning model to understand the context and intensity of the text. Figure \ref{fig:Figure 2} shows our proposed rumour classification model using BERT's sentence embedding. 

The proposed model consists of four stages.  The first stage collects text data from a dataset. In this study, we used \emph{PHEME dataset} which consists of a collection of tweets classified as rumour or non-rumour. Before proceeding to the second stage, we tokenized each tweet,  resulting in a sequence of tokens that represents each tweet's sentence. The second stage is sentence embedding using BERT\textsubscript{BASE} where the sequence of tokens from the previous stage is the input. We utilised Sentence-Transformer library \cite{SBERT} to embed a sequence of tokens from each tweet into a vector. This vector is a numeric form representing each tweet's sentence so that the proposed model can understand and perform  mathematical operations on it. We used these embeddings to train the text classification model for predicting rumours in the third stage. We evaluated the prediction results in the fourth stage.  

\subsection{BERT (Bidirectional Encoder Representation from Transformer)}

BERT is a new language representation model which is designed as a deep bidirectional transformer that captures information from the unlabeled text by joining the representations from both left and right of a token's context from all layers \cite{devlin2018bert}.  BERT understands word relationships in a bidirectional way and then generates a representation vector for each word based on the word's relationships in a sentence. 

\begin{figure}[hb]
    \begin{center}
    \centering
    \includegraphics[scale=0.9]{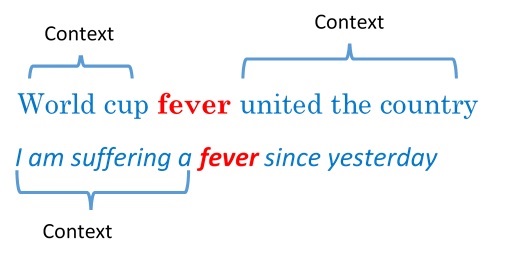}
    \caption{How BERT captures the context from both the left and right side of a word} 
     \label{FIG:Figure3}
     \end{center}
\end{figure}

 Figure \ref{FIG:Figure3} presents an example to illustrate the bidirectional model in two sentences: \emph{World cup fever united the country} and \emph{I am suffering a fever since yesterday}. Only considering either the left or the right context will result in a wrong representation of the meaning of the word `fever'. BERT considers both the left and the right context of the word `fever' before making a representation. From these two sentences, BERT considers \emph{world cup} and \emph{united the country} to represent the word `fever' in the first sentence and \emph{I am suffering} and \emph{since yesterday} to represent the word `fever' in the second sentence. Hence, the word `fever' in the sentences \emph{World cup fever united the country} and \emph{I am suffering a fever since yesterday} will have  different vector representations.

\subsubsection{BERT architecture model}
As mentioned by its name, BERT is a bidirectional transformer. A primary transformer consists of an encoder to read the text input and a decoder to predict the task's output.  BERT only needs the encoder part since BERT's goal is to generate a language representation model.  BERT's encoder input is a sequence of tokens that is converted into a vector. BERT provides two architecture models: BERT\textsubscript{BASE} and BERT\textsubscript{LARGE}.  BERT\textsubscript{BASE} has 12 encoder layers, 768 hidden layers and 110 millon parameters whereas BERT\textsubscript{LARGE} has 24 encoder layers, 1024 hidden layers and 340 million parameters \cite{devlin2018bert}. This study uses BERT\textsubscript{BASE} to reduce the computational complexity. BERT\textsubscript{BASE} represents each tweet's sentence into 1x768 vector.

\subsubsection{BERT Input/Output Representation}

\begin{figure*}[ht]
    \centering
        \includegraphics[scale=0.8]{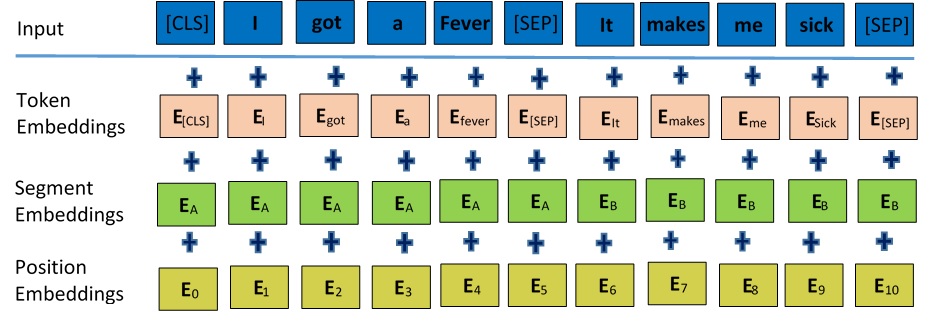}
    \caption{The input representation for BERT  \cite{devlin2018bert}. [CLS] is a special symbol added in the beginning, and [SEP] is a special separator token added at each sentence's end.}
    \label{fig:Figure 4}
\end{figure*}

BERT is able to represent both a single sentence and a set of sentences in one token sequence. We borrowed a figure (Figure \ref{fig:Figure 4}) and terms from Devlin et al. \cite{devlin2018bert} as an inspiration to explain BERT input representation in more details. In BERT, a `sentence' refers to a continuous text instead of a sentence in a linguistic sense. A `sequence' refers to the sequence of tokens from either a single sentence or a set of sentences grouped together for BERT’s input.  

Consider two sentences:``I got a fever” as sentence A and ``It makes me sick” as sentence B. It can be observed in Figure \ref{fig:Figure 4}, BERT's input representation is the sum of the token, segment and position embeddings. They are explained below.  
\begin{enumerate}
    \item \emph{Token Embeddings} are tokens added in the beginning of each sentence [CLS] and at the end of each sentence [SEP]. 
    \item \emph{Segment Embeddings} are the tokens that are added as a marker to distinguish sentences.  BERT distinguishes the sentences in two ways by separating them by a special token [SEP] and adding an additional token to indicate whether it belongs to sentence A or sentence B. BERT denotes the input embedding as \emph{E}, so that all the tokens for sentence A are marked as E\textsubscript{A}, and for sentence B marked as E\textsubscript{B}. Segment embedding enables BERT to take the sentence pairs and pack together into a single sequence.
    \item \emph{Position Embeddings} are the tokens added to indicate the position of each token in the sentence. BERT learns and adds a positional embedding token and utilises it to express the position of words in a sentence.  Those positional embedding tokens are marked as $E_0$ to $E_n$, where n represents the sequence number of each token. 
\end{enumerate}

\subsection{Classifiers to identify rumours}

We used BERT to represent each tweet's sentence into a numerical vector. Then we used all the vectors to train a text classification model using various supervised learning approaches to classify whether a tweet is a rumour or not. We utilised some techniques reported in the literature for achieving good performance on text classification from various studies. The selected classification techniques are Support Vector Machines (SVM) \cite{Hassan2018,vijeev2018hybrid,kotteti2018multiple} Logistic Regression (LR) \cite{Hassan2018,vijeev2018hybrid,kotteti2018multiple}, Naïve Bayes Classifier (NBC) \cite{Hassan2018,vijeev2018hybrid}, AdaBoost \cite{wang2017ecnu,nasir2021fake}, and K-Nearest Neighbors (KNN) \cite{Hassan2018,vijeev2018hybrid}. Additionally, we also used BERT vectors to train the classifier model based on Multilayer Perceptron (MLP). MLP is a deep artificial neural network that consists of more than one perceptron. A MLP comprises an input layer to receive the signal, an output layer to predict the input, and an arbitrary number of hidden layers between the input and output layers \cite{Taud2018}.

\subsection{Evaluation Method}
We used a confusion matrix which comprises: true positive (TP), true negative (TN), false positive (FP) and false-negative (FN). They are explained as:
\begin{itemize}
   \item \textit{True-positive (TP)} are non-rumour tweets that are correctly predicted as non-rumour tweets.
   \item \textit{False-negative (FN)} are non-rumour tweets that are incorrectly predicted as rumour tweets.
   \item \textit{False-positive (FP)} are rumour tweets that are incorrectly predicted as  non-rumour tweets.
   \item \textit{True-negative (TN)} are rumour tweets that are correctly predicted as rumour tweets.
\end{itemize}
By using the confusion matrix, we evaluated the performance of our model by calculating Accuracy, Precision, Recall and F1-score using the following equations.
\begin{equation}
Accuracy=\frac{TP+TN}{TP+TN+FP+FN}\\
\end{equation}

\begin{equation}
Precision (P)=\frac{TP}{TP+FP}\\
\end{equation}

\begin{equation}
Recall (R)=\frac{TP}{TP+FN}\\   
\end{equation}

\begin{equation}
F1 =\frac{2(P)(R)}{P+R}\\    
\end{equation}
\\

\section{Experiments}

\begin{table*}[htb]
\caption{State-of-the-art rumour detection models for PHEME dataset (best results for each parameter are mentioned in bold). }
\centering
\small
\begin{tabular} {|p{3cm}|p{3.5cm}|p{1cm}|p{1.3cm}|p{1.5cm}|p{1.5cm}|} 
\hline
{Previous works} & 
{Method} &
\multicolumn{4}{|c|}{Best Performance}\\
\cline{3-6} 
&  & Accuracy & Precision & Recall & F1-score \\
\hline
Zubiaga et al.\cite{Zubiaga2016} & 
Conditional Random field (CRF) based on content and social features  & 
- & 0.667 & 0.556 & 0.607\\
\hline 
Hasan and Haggag\cite{Hassan2018} &  
Various supervised learning algorithms & 
0.784 &  0.796  & \textbf{0.919}  & 0.852 \\
\hline 
Ajao et al.\cite{Ajao2018} &  
Combining CNN and LSTM models & 0.823  & 0.443  & - & -\\
\hline 
Kotteti et al.\cite{kotteti2018multiple} &  Using time series data to reduce time and  supervised learning algorithms  & - & \textbf{0.949} & 0.356 & 0.518\\
\hline 
Alkhodair et.al\cite{Alkhodair2020} & 
Using word embedding and CNN & 
- & 0.728-R, 0.833-NR & 0.706-R, 0.847-NR & 0.716-R,  0.839-NR, \textbf{0.795-all} \\
\hline 
Zhou and Li \cite{Zhou2019} & 
Reinforcement learning  &
\textbf{0.858} & 0.843 & 0.735 & 0.785\\
\hline 
Xu et al. \cite{Xu2020} & 
Topic-driven rumor detection (TDRD), by combining topic model and CNN & 
0.8266 & \textbf{0.813-R}, 0.831-NR & 
0.6355-R, \textbf{0.9249-NR} & 
0.7120-R, 0.8755-NR\\
\hline 
Bharti and Jindal \cite{Bharti2021} &  
Using some machine learning techniques and CNN & - & 0.790-R    \textbf{0.870-NR} & \textbf{0.760-R},   0.890-NR & 
\textbf{0.770-R},   \textbf{0.880-NR}\\
\hline
\multicolumn{6}{l}{Abbreviations: R:Rumours, NR: Non-Rumours, all: All class}
\end{tabular}
\label{table3}
\end{table*}

\subsection{Datasets}
Collection of data for classification of rumour is a challenging task and there are only a few  publicly available datasets \cite{Bondielli2019}.  We used the dataset from PHEME project \cite{Zubiaga2016} which is publicly available and considered as a benchmark dataset for this problem. The dataset contains 1969 tweets labelled as  rumours and 3822 tweets labelled as non-rumours. We split each class of the dataset into two parts: 70\% for training and 30\% for testing. 

We used the five-fold cross validation technique to validate the results from the selected classifiers. We randomly divided the datasets into five groups of folds of approximately equal size. A unique group was treated as a validation set for the four other groups, which were treated as training dataset.  

\subsection{Baseline and state of the art}

Table \ref{table3} summarises the state-of-the-art methods and their performances on PHEME dataset. We used those studies as the baseline to compare our proposed model's results.

\subsection{Experimental Steps}
\begin{figure}[ht]
    \centering
    \includegraphics[scale=0.6]{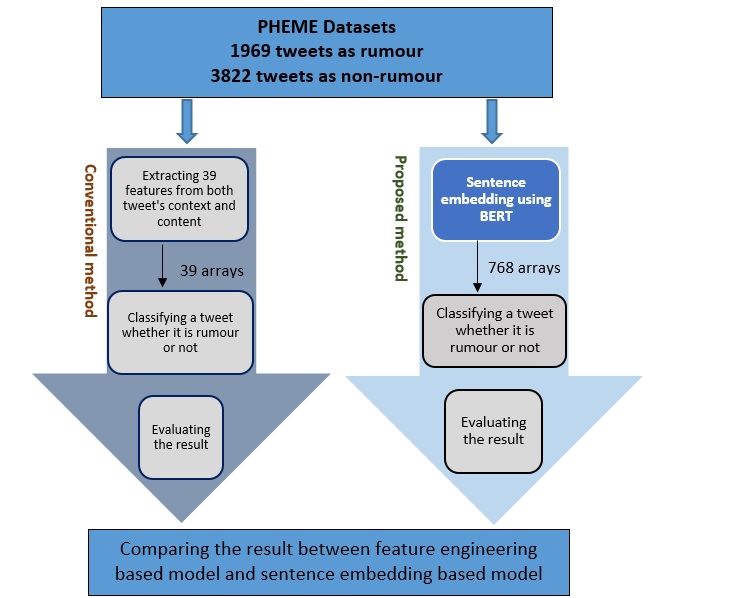}
    \caption{The experimental steps for rumour detection using feature extraction and BERT.}
    \label{fig:Figure 5}
\end{figure}

We experimented with both the conventional method using 39 features mentioned in Figure \ref{fig:Figure 1} and the proposed method using BERT's vectors as mentioned in Figure \ref{fig:Figure 2}.  We then compared the performance of these two methods on rumour detection.  Figure \ref{fig:Figure 5} illustrates the comparison procedure for both experimental procedures to classify rumour and non-rumour tweets. 

For the proposed method using BERT, we preprocessed and tokenised the tweets and concatenated the tokens to obtain the tokenised version of each tweet's sentence. Then we encoded the tokenised sentences into vectors using the SentenceTransformer library  to get 1x768 arrays representing each tweet. We saved these vectors as a new dataset. 

We used these vectors to train a classification model using various standard supervised learning algorithms as a baseline, including Support Vector Machines (SVM), Logistic Regression (LR), Naïve Bayes Classifier (NBC), AdaBoost and K-Nearest Neighbors (KNN). We also used the vectors to train a classification model using a four-layer MLP (4L-MLP). 
We preferred 4L-MLP over RNN or CNN which are widely used for classification tasks due to its simplicity and lower computational complexity. RNNs are more suitable for time series or sequential data \cite{minaee2020deep} and CNNs for images. 
 
We also utilised the feature extraction approach by extracting 39 features from the tweets' contexts and contents based on methods proposed in previous studies, as presented in Table \ref{table1} and Table \ref{table2}. We converted the values of these features into integer data type to obtain 39 arrays of features for each tweet. We used those arrays to train the classification models using the same approaches as used for the BERT-based proposed model. We then compared the performance of these two approaches to find the best model to detect rumours.
\vspace{-1pt}
We experimented with all the classifiers as mentioned in the earlier sections. We selected the top two classifiers that achieved the highest accuracy and thoroughly evaluated them using 5-fold cross validation. This helped to estimate the true prediction error of models by re-sampling data \cite{berrar2019cross}.
For the best classifier, we applied L2 regularisation and dropout techniques to prevent over-fitting. Regularisation is a method to reduce the generalisation error but not the training error by modifying the cost function \cite{Goodfellow2015}. L2 regularisation  minimises the  squared magnitude of weights to reduce generalisation error. Similarly, dropout is a technique to address the over-fitting problem by modifying the network itself. It drops connection units  from the neural network randomly during training  to ensure that the model is not over-fitted with the data \cite{srivastava2014dropout}.

\subsection{Results and discussion}

As shown in Table \ref{table4}, the confusion matrix summarises each classifier model's prediction results. Table \ref{table4} shows the true positive, false positive, true negative, and false negative values for each classification model's predictions. These values were used in equations (1) – (4) to measure accuracy, precision, recall, and F1-score.

\begin{table}[ht]
\centering
\caption{Confusion matrix of each classifier prediction result on rumour detection using feature extraction and BERT on PHEME datasets}
\begin{tabular} 
{|p{1cm}|p{1.37cm}|p{1.5cm}|p{1cm}|p{1cm}|}
\hline
Classifier Model & Data model & 
Prediction & Non-Rumours & 
Rumours \\
\hline
\multirow{2}{4em}{Support Vector Machine} & \multirow{2}{4em}{BERT} & Non-Rumours & 1000 & 155 \\ 
& & Rumours & 160 & 422 \\ 
\cline {2-5}
& \multirow{2}{4em}{39 Features}& Non-Rumours & 1035 & 332 \\ 
& & Rumours & 125 & 245 \\
\hline
\multirow{2}{4em}{Logistic Regression} & \multirow{2}{4em}{BERT} & Non-Rumours & 1020 & 154 \\ 
& & Rumours & 140 & 423 \\ 
\cline {2-5}
& \multirow{2}{4em}{39 Features}& Non-Rumours & 1037 & 332 \\ 
& & Rumours & 123 & 245 \\
\hline
\multirow{2}{4em}{Naive Bayes} & \multirow{2}{4em}{BERT} & Non-Rumours & 835 & 131 \\ 
& & Rumours & 325 & 446 \\ 
\cline {2-5}
& \multirow{2}{4em}{39 Features}& Non-Rumours & 645 & 129 \\ 
& & Rumours & 515 & 448 \\
\hline
\multirow{2}{4em}{ADA Boost} &  \multirow{2}{4em}{BERT} & Non-Rumours & 983 & 198 \\ 
& & Rumours & 177 & 379 \\ 
\cline {2-5}
& \multirow{2}{4em}{39 Features} & Non-Rumours & 
1001 & 298 \\ 
& & Rumours & 159 & 281 \\
\hline
\multirow{2}{4em}{K-Nearest Neighbor} & \multirow{2}{4em}{BERT} & Non-Rumours & 989 & 108 \\ 
& & Rumours & 171 & 469 \\ 
\cline {2-5}
& \multirow{2}{4em}{39 Features}& Non-Rumours & 
914 & 260 \\ 
& & Rumours & 246 & 317 \\
\hline
\multirow{2}{4em}{4-Layers MLP} & \multirow{2}{4em}{BERT} & Non-Rumours & 1016 & 125 \\ 
& & Rumours & 144 & 452 \\ 
\cline {2-5}
& \multirow{2}{4em}{39 Features}& Non-Rumours & 972 & 237 \\ 
& & Rumours & 188 & 340 \\
\hline
\end{tabular}
\label{table4}
\vspace{-2mm}
\end{table}

Table \ref{table5} compares the accuracy, precision, recall, and F1-score for each classifier. Except for the recall of SVM, LR and ADA-Boost models for the non-rumour class and the recall of Naive Bayes classifier model for the rumour class, we found that the BERT-based classifier model outperformed the feature-based classifier model.
BERT-based classifier model achieved accuracy, precision, recall, and F1-score around 10\%  higher than those of feature-based classifier models for almost all techniques and  for all classes.  For the non-rumour class, the BERT-based classifier model using SVM, LR and ADA-Boost demonstrated slightly different results. BERT-based classifier model using those techniques obtained recall of around 3\%, 1.5\%, and 1.6\% higher for SVM, LR and ADA Boost respectively, which were slightly lower than the feature-based classifier models using the same techniques. The BERT-based Naive Bayes classifier model achieved lower recall for the rumour class than the feature-based Naive Bayes classifier model by about 0.3\% (see Table \ref{table5}). Based on these findings, we were confident that sentence embedding with BERT was a promising approach for identifying rumour tweets without extracting any features. 

We then moved on to the next step to determine the best rumour detection model to improve the current state of the art results. To find the best model for rumour detection, we chose two models which showed the best performance. As shown in Table  \ref{table5}, BERT-based K-NN, and 4L-MLP models performed the best, with accuracies of 0.839 and 0.845 respectively, and precision of 0.817 and 0.824 respectively, for all class predictions. For this reason, we only selected the BERT-based classifier model using K-NN and 4L-MLP  to be validated using the five-fold cross-validation technique.
\par To perform five-fold cross-validation, we set the learning rate to 0.0002, the batch size to 512, and Adam as the optimiser to complete  re-sampling procedure for each batch within 20 minutes of training time. As shown in Table \ref{table6}, the BERT-based classifier model using 4L-MLP consistently outperformed the BERT-based classifier model using K-NN. Hence, we concluded that the BERT-based classifier model using 4L-MLP is the best model to detect rumours on Twitter. 

\begin{center}
\begin{table*}[ht]
\centering
\caption{Experimental results of rumour classification models on PHEME datasets both using BERT and feature engineering}
\begin{tabular} {*{12}{|c}|}
\hline
\multirow{2}{5em}{Classifier Model} & \multirow{2}{4em}{Data Model} & \multicolumn{4}{|c|}{All Class} & \multicolumn{3}{|c|}{Non-Rumours} & \multicolumn{3}{|c|}{Rumours}\\ 
\cline{3-12}
& & Acc & Prec & Recall & F1 & Prec & Recall & F1 & Prec & Recall & F1 \\ 
\hline
\multirow{3}{5em}{Support Vector Machine} & {BERT} & 0.819 & 0.795 & 0.797 & 0.796 & 0.866 & 0.862 & 0.864 & 0.725 & 0.731 & 0.728 \\ 
& {39 Features} & 0.737 & 0.710 & 0.658 & 0.683 & 0.757 & \underline{0.892} & 0.819 & 0.662 & 0.425 & 0.517 \\ 
& {Improvement} & \emph{0.082}	& \emph{0.086} & \emph{0.138} & \emph{0.113}	 & \emph{0.109} &	\emph{-0.030} &	\emph{0.045} &	\emph{0.063} &\emph{0.307} &	\emph{0.211} \\ 
\hline
\multirow{3}{5em}{Logistic Regression} & {BERT} & 0.831 &	0.810 & 0.806 &	0.808 &	0.869 &	0.879 &	0.874 &	0.751 &	0.733 &	0.742\\
& {39 Features} & 0.738 &	0.712 &	0.659 &	0.684 &	0.757 &	\underline{0.894} &	0.820 &	0.666 &	0.425 &	0.519 \\ 
    & {Improvement} & \emph{0.093} &	\emph{0.098} &	\emph{0.147} &	\emph{0.124} &	\emph{0.111} &	\emph{-0.015} &	\emph{0.054} &	\emph{0.086} &	\emph{0.308} &	\emph{0.224} \\ 
\hline
\multirow{3}{5em}{Naive Bayes} & {BERT} & 0.737 &	0.721 &	0.746 &	0.734 &	0.864 &	0.720 &	0.786 &	0.578 &	0.773 &	0.662\\
& {39 Features} & 0.629 &	0.649 &	0.666 &	0.658 &	0.833 &	0.556 &	0.667 &	0.465 &	\underline{0.776} &	0.582 \\ 
& {Improvement} & \emph{0.108} &	\emph{0.072} &	\emph{0.080} &	\emph{0.076} &	\emph{0.031} &	\emph{0.164}	& \emph{0.119} &	\emph{0.113} &	\emph{-0.003} &	\emph{0.080} \\ 
\hline
\multirow{3}{5em}{ADA Boost} & {BERT} & 0.784 &	0.757 &	0.752 &	0.755 &	0.832 &	0.847 &	0.840 &	0.682 &	0.657 &	0.669\\
& {39 Features} & 0.738 &	0.705 &	0.675 &	0.690 &	0.772 &	\underline{0.863} &	0.815 &	0.639 &	0.487 &	0.553 \\ 
& {Improvement} & \emph{0.046} &	\emph{0.052} &	\emph{0.077} &	\emph{0.065} &	\emph{0.061} &	\emph{-0.016} &	\emph{0.025} &	\emph{0.043} &	\emph{0.170} &	
\emph{0.116} \\ 
\hline
\multirow{3}{5em}{K-Nearest Neighbor} & {BERT} & \textbf{0.839} &	\textbf{0.817} &	\textbf{0.833} &	\textbf{0.825} &	\textbf{0.902} &	\textbf{0.853} &	\textbf{0.876} &	\textbf{0.733} &	\textbf{0.813} &	\textbf{0.771}\\
& {39 Features} & 0.709 &	0.671 &	0.669 &	0.670 &	0.779 &	0.788 &	0.783 &	0.563 &	0.549 &	0.556 \\ 
& {Improvement} & \emph{0.131} & \emph{0.146} &	\emph{0.164} &	\emph{0.155} &	\emph{0.123} &	\emph{0.065} &	\emph{0.093} &	\emph{0.170} &	\emph{0.263} &	\emph{0.215} \\ 
\hline
\multirow{3}{5em}{4-Layers MLP } & {BERT} & \textbf{0.845} &	\textbf{0.824} &	\textbf{0.830} &	\textbf{0.827} &	\textbf{0.890} &	\textbf{0.876} &	\textbf{0.883} &	\textbf{0.758} &	\textbf{0.783} &	\textbf{0.771}\\
& {39 Features} & 0.755 &	0.724 &	0.714 &	0.719 &	0.804 &	0.838 &	0.821 &	0.644 &	0.589 &	0.615 \\ 
& {Improvement} & \emph{0.090} &	\emph{0.100} &	\emph{0.116} &	\emph{0.108} &	\emph{0.086} &	\emph{0.038} &	\emph{0.062} &	\emph{0.114} &	\emph{0.194} &	\emph{0.155} \\ 
\hline
\multicolumn{12}{l}{The improvement performance between features-based and BERT-based model are presented in italic.}\\
\multicolumn{12}{l}{The two best model's performance  for each parameter are mentioned in bold.}\\
\multicolumn{12}{l}{The exception pattern where the feature-based model outperform the BERT-based model are illustrated in underlined}
\end{tabular}
\label{table5}
\end{table*}
\end{center}

\begin{center}
\begin{table*}[ht]
\centering
\caption{Performance comparison between K-NN and 4L-MLP model using BERT to classify rumours on the five-fold cross-validation}
\begin{tabular} {*{12}{|c}|} 
\hline
\multirow{2}{4em}{Classifier Model} & \multirow{2}{4em}{Fold} & \multicolumn{4}{|c|}{All Class} & \multicolumn{3}{|c|}{Non-Rumours} & \multicolumn{3}{|c|}{Rumours}\\ 
\cline{3-12}
& & Acc & Prec & Recall & F1 & Prec & Recall & F1 & Prec & Recall & F1 \\ 
\hline
\hline
\multirow{6}{4em}{K-Nearest Neighbor} & {Fold-1} & 0.814 &	0.794 &	0.801 &	0.797 &	0.869&	0.844&	0.857&	0.718&	0.757&	0.737\\
& {Fold-2} & 0.829&	0.811&	0.823&	0.817&	0.889&	0.843&	0.866&	0.733&	0.803&	0.766 \\ 
& {Fold-3} & 0.831&	0.812&	0.818&	0.815&	0.879&	0.861&	0.870&	0.745&	0.774&	0.759 \\ 
& {Fold-4} & 0.809& 	0.782& 	0.800& 	0.791& 	0.886& 	0.825& 	0.855& 	0.678& 	0.775& 	0.723 \\ 
& {Fold-5} & 0.814 &	0.793 &	0.805 &	0.799 &	0.878 &	0.832 &	0.854 &	0.708 &	0.779 &	0.742 \\ 
& \emph{Average} & \emph{0.820} &	\emph{0.798} &	\emph{0.809} &	\emph{0.804} &	\emph{0.880} &	\emph{0.841} &	\emph{0.860} &	\emph{0.716} &	\emph{0.778} &	\emph{0.745} \\
\hline
\multirow{6}{4em}{4-Layers MLP} & {Fold-1} & 0.846 &	0.827 &	0.838 &	0.832 &	0.898 &	0.863 &	0.880 &	0.757 &	0.812 &	0.783\\
& {Fold-2} & 0.846 &	0.836 &	0.820 &	0.828 &	0.864 &	\underline{0.907} &	0.885 &	\underline{0.808} &	0.734 &	0.769 \\ 
& {Fold-3} & \underline{0.856} &	\underline{0.839} &	\underline{0.846} &	\underline{0.843} &	\underline{0.901} &	0.878 &	\underline{0.889} &	0.777 &	\underline{0.815} &	\underline{0.796} \\ 
& {Fold-4} & 0.832 &	0.813 &	0.791 &	0.802 &	0.856 &	0.904 &	0.879 &	0.770 &	0.679 &	0.722 \\ 
& {Fold-5} & 0.847 &	0.829 &	0.839 &	0.834 &	0.898 &	0.866 &	0.882 &	0.760 &	0.812 &	0.785 \\ 
& \textbf{Average} & \textbf{0.846} &	\textbf{0.829} &	\textbf{0.827} &	\textbf{0.828} &	\textbf{0.883} &	\textbf{0.884} & \textbf{0.883} &	\textbf{0.774} &	\textbf{0.770} &	\textbf{0.771} \\
\hline
\multicolumn{12}{l}{The average result for each parameter of KNN model are presented in italic.}\\
\multicolumn{12}{l}{The best result for each parameter are presented in underline.}\\
\multicolumn{12}{l}{The average result for each parameter of 4L-MLP model are presented in italic.}\\

\end{tabular}
\label{table6}
\end{table*}
\end{center}

\begin{center}
\begin{table*}[ht]
\centering
\caption{Performance comparison between 4L-MLP model with regularisation and dropouts.}
\begin{tabular} {*{11}{|c}|}
\hline
\multirow{2}{2pt}{Classifier Model} &  \multicolumn{4}{|c|}{All} & \multicolumn{3}{|c|}{Non-Rumours} & \multicolumn{3}{|c|}{Rumours}\\ \cline{2-11}
 & Acc & Prec & Recall & F1 & Prec & Recall & F1 & Prec & Recall & F1 \\ 
\hline
{4L-MLP } & 0.846 &	0.829 &	0.827 &	0.828 &	0.883 &	0.884 &	0.883 &	0.774 &	0.770 &	0.771 \\
\hline
{4L-MLP, Reg and Dropout} &  \emph{0.869} &	\emph{0.855} &	\emph{0.848} &	\emph{0.852} &	\emph{0.895} &	\emph{0.911} &	\emph{0.903} &	\emph{0.815} &	\emph{0.785} &	\emph{0.799} \\
\hline
{Improvement} &  \textbf{0.023} &	\textbf{0.026} & \textbf{0.021} &	\textbf{0.024} & \textbf{0.012} &	\textbf{0.027} & \textbf{0.02} &	\textbf{0.04} & \textbf{0.015} &	\textbf{0.029} \\
\hline
\multicolumn{11}{l}{The improved results for each parameter are presented in bold.}
\end{tabular}
\label{table7}
\vspace{-3mm}
\end{table*}
\end{center}

\vspace{-20mm}
We applied L2-regularisation and dropout techniques to prevent over-fitting on our best model by setting the dropout probability at 0.5, the weight decay at 0.00001, the batch size at 512, and  Adam as the optimiser. We achieved performance improvement on an average at around 2.37\% for all parameters. The details of the improvements are shown in bold in Table \ref{table7}. For all classes, we achieved an accuracy of 0.869, precision of 0.855, recall of 0.848, and F1-score of 0.852.  We obtained a precision of 0.895, a recall of 0.911, and an F1-score of 0.903 for the non-rumour class and a precision of 0.815, recall of  0.785,  and an F1-score of 0.799  for the rumour class (See Table \ref{table7}).

\begin{table*}[ht]
\centering
\caption{Performance comparison between the proposed model and previous state-of-the-arts}

\begin{tabular} {*{11}{|c}|}
\hline
\multirow{2}{4em}{Previous works} &  \multicolumn{4}{|c|}{All} & \multicolumn{3}{|c|}{Non-Rumours} & \multicolumn{3}{|c|}{Rumours}\\ \cline{2-11}
 & Acc & Prec & Recall & F1 & Prec & Recall & F1 & Prec & Recall & F1 \\ 
\hline
{Zubiaga et al. \cite{Zubiaga2016} } & - & 0.667 & 0.556 & 0.607 &	- &	- &	- &	- &	- &	- \\
\hline
{Hassan and Haggag \cite{Hassan2018}} & 0.784 &  0.796  & \underline{0.919}  & \underline{0.852} &	- &	- &	- &	- & - &	- \\
\hline
{Ajao et al. \cite{Ajao2018}} & 0.8229  & 0.4435  & - & - & - &	- & - &	- & - & -  \\
\hline
{Kotteti et al. \cite{kotteti2018multiple}} & -  & \underline{0.949}  & 0.356  & 0.518 & - &	- & - &	- & - & -  \\
\hline
{Zhou and Li \cite{Zhou2019}}  & \underline{0.858} & 0.843 & 0.735 & 0.785 & - &	- & - &	- & - & -  \\
\hline
{Alkhodair et al. \cite{Alkhodair2020}} & - & - & - & 0.795 & \underline{0.833} & 0.847 &  0.839 & 0.728 & 0.706 & 0.716 \\
\hline
{Xu et al. \cite{Xu2020}} & 0.8266 & - & - & - & 0.831 & \underline{0.9249}  &  0.8755 & \underline{0.813} & 0.6355 & 0.7120\\
\hline
{Bharti and jindal \cite{Bharti2021}} & - & - & - & - & 0.87 & 0.89  &  \underline{0.88} & 0.79 & \underline{0.76} & \underline{0.77}\\
\hline
{BERT and 4L-MLP} & \textbf{0.869} & \textbf{0.855} & \textbf{0.848} & \textbf{0.852} & \textbf{0.895} &  \textbf{0.911} & \textbf{0.903} & \textbf{0.815} & \textbf{0.785} &  \textbf{0.799}  \\
\hline
\multicolumn{11}{l}{The proposed model's performance are presented in bold.}\\
\multicolumn{11}{l}{The best previous studies performance's result for each parameter of each class  are mentioned in underline.}
\end{tabular}
\label{table8}
\end{table*}

We then compared our best model (BERT-based classifier using 4L-MLP) with the existing state of the art models from literature on the PHEME dataset, as shown in Table \ref{table8}. Our proposed model outperforms the state of the art techniques by achieving an accuracy of 0.869.  Kotteti {em et. al.} \cite{kotteti2018multiple} achieved the highest precision of 0.949, while our model achieves 0.855 precision and is the second-highest among all classes. For the rumour and non-rumour classes, our model achieves the best precision  0.815 and 0.895, respectively.   
Hassan {\em et al.} \cite{Hassan2018} achieved the highest recall of 0.919  for all classes. For the rumour class, our model outperformed the other models at 0.785 of recall.  For the non-rumour class, we achieved a recall of 0.911,  slightly lower than the $0.9249$ reported in \cite{xu2020near}. 
Hassan {\em et al.} \cite{Hassan2018} reported the best F-1 score at 0.852, which is same as our model, 0.852 for all classes. For the rumour and non-rumour classes, our model outperformed the other models  by obtaining the F1-scores of 0.799 and 0.903, respectively. 
Based on the comparisons,  our proposed model is better than the other models in overall performance. Table \ref{table8} shows that our proposed model outperforms the earlier  models  by achieving the accuracy of 0.869, precision of 0.815 and 0.895 for rumour and non-rumour classes respectively.  Our model attained recall scores of 0.785 and 0.911 for rumour and non-rumour classes respectively and obtained F1-scores of 0.799 for the rumour class, and 0.903  for the non-rumour class.

\section{Conclusion}

Though social networks have opened up unprecedented opportunities for expressing opinions, they are fraught with the danger of spreading rumours and false information. It is important to detect and purge rumours for these platforms as fast as possible, and that is only possible with automatic detection of rumours due to the sheer volume of posts. We have addressed the problem of automatic rumour detection in tweets. 

The majority of rumour detection studies rely on the feature extraction process, which is time-consuming. Recently, Google introduced BERT, a novel transformer to represent language. BERT can capture and represent the contextual meaning of a sentence into numeric arrays to enable a model to understand and perform mathematical operations. In this study, we examined whether BERT's output can be used to train a rumour detection model.  We also proposed a novel approach by leveraging BERT's sentence embedding and the text of tweets  to identify rumours. 

Our experimental results showed that BERT's sentence embedding could be used to distinguish rumour and non-rumour tweets, without extracting tweets' features. By utilising BERT's sentence embedding, various supervised classification models demonstrated better performance results compared to feature-based classification models. Furthermore, by leveraging BERT's sentence embedding-based classification model using 4L-MLP technique we have presented a new state-of-the-art  rumour detection model for Twitter by obtaining 0.869  accuracy, 0.855 precision, 0.848  recall and 0.852 F1 score. We hypothesize that larger datasets of tweets containing rumour and non-rumour labels can further improve these results.

\vspace{3mm}
\section*{Acknowledgment}

This paper and research behind it would not have been possible without the exceptional support of Sebelas Maret University as the sponsor.  We would like to thank to Sebelas Maret University that has provided not only financial support but also facilities and moral support to resume my research during the critical time of COVID-19 pandemic.


\bibliographystyle{IEEEtran}
\bibliography{IEEEbert}
 
\vspace{-10mm}
\begin{IEEEbiography}
[{\includegraphics
[width=1in,height=1.25in,clip,keepaspectratio]
{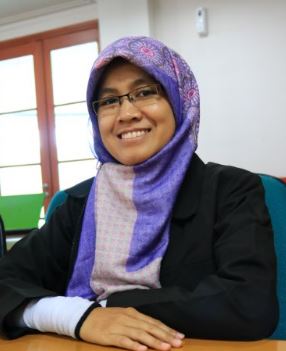}}]{Rini Anggrainingsih}

She received her bachelor degree from Diponegoro University and received her master degree from Gadjahmada University in Indonesia. She is currently working as academic staff at Sebelas Maret University and has been pursuing her PhD at The University of Western Australia. Her current research interest are Twitter data credibility analysis using machine learning approaches to improve info-surveillance on social media.
\end{IEEEbiography}

\begin{IEEEbiography}
[{\includegraphics
[width=1in,height=1.25in,clip,keepaspectratio]
{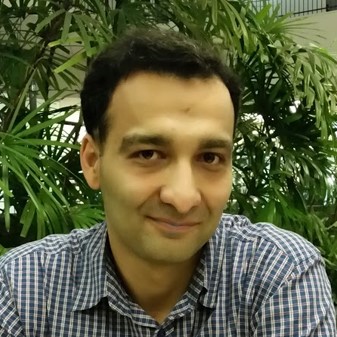}
}]{Ghulam Mubashar Hassan}

Dr. Ghulam Mubashar Hassan received his B.Sc. Electrical and Electronics Engineering degree (with honors) from University of Engineering and Technology (UET) Peshawar, Pakistan. He completed his MS in Electrical and Computer Engineering from Oklahoma State University USA and his PhD from The University of Western Australia (UWA) in Joint Schools of Computer Science \& Software Engineering and Civil \& Resource Engineering. He was valedictorian and received many awards for his PhD. Currently, he is working in UWA and previously he worked in UET Peshawar and King Saud University. His research interests are multidisciplinary problems which include using artificial intelligence, machine learning, pattern recognition, optimization in different fields of engineering and education.
\end{IEEEbiography}

\vspace{-130mm}
\begin{IEEEbiography}
[{\includegraphics
[width=1in,height=1.25in,clip,keepaspectratio]
{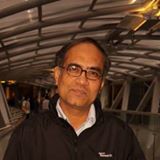}
}]{Amitava Datta}

Amitava Datta (A’04–M’10) received the M.Tech. and Ph.D. degrees from IIT Madras in 1988 and 1992, respectively. He did the post-doctoral research at the Max Planck Institut für Informatik, Germany, and the University of Freiburg, Germany. He joined the University of New England in 1995 and The University of Western Australia in 1998, where he is currently a Professor with the School of Computer Science and Software Engineering. His current research interests are in optical computing, data mining, bioinformatics, and social network analysis. He has authored over 150 papers in various international journals and conference proceedings, including the IEEE Transactions on Computers, the IEEE Transactions on Parallel and Distributed Systems, the IEEE Transactions on Visualization and Computer Graphics, the IEEE Transactions on Mobile Computing, the IEEE/ACM Transactions on Networking, the IEEE Transactions on Computational Social Systems and the IEEE Transactions on Systems, Man, and Cybernetics.
\end{IEEEbiography}







\end{document}